\documentclass[twocolumn,10pt]{asme2ej}

\usepackage{graphicx}
\usepackage{amsmath}
\usepackage{hyperref}   
\hypersetup{
	colorlinks=true,
	linkcolor=blue,
	citecolor=blue,
	urlcolor=blue,
}
\usepackage[square,numbers]{natbib}

\title{Wireless Self-Powered Visual and NDE low-Cost Inspection System For Small Diameter Live Gas Distribution Mains}

\author{Shivani Naik, Arjun Kumar, Nitinesh Yadav and K. M. Santosh}

\begin{document}

\maketitle    

The arrangement of an in-pipe climbing robot that works using a sharp transmission part to explore complex relationship of lines. Standard wheeled/continued in-pipe climbing robots are leaned to slip and take while researching in pipe turns. The instrument helps in achieving the really unavoidable consequence of getting out slip and drag in the robot tracks during progression. The proposed transmission likes the useful uttermost scopes of the standard two-yield transmission, which is fostered the fundamental time for a transmission with three outcomes. The instrument decisively changes the track velocities of the robot considering the powers applied on each track inside the line relationship, by getting out the fundamental for any wonderful control. The entertainment of the robot crossing in the line network in different direction and in pipe-turns without slip shows the proposed course of action's ampleness.

\section{Introduction}

Pipe networks are unavoidable, from an overall perspective used to move liquids and gases in sties and metropolitan associations. Most often, the lines are covered to concur with the security rules and to avoid hazards. This makes audit and upkeep of lines really testing. Covered lines are particularly organized to frustrating, use, scale outline, and break initiation, achieving movements or damages that may influence appalling episodes. Different Inspection Robots \cite{adriansyah2017optimization} were proposed in the past to lead standard preventive evaluations to avoid disasters. In addition, bio-vivified robots with crawler, inchworm, walking parts \cite{ravina2010low}, and screw-drive\cite{jalal2022pipe} systems were in like way shown to be fitting for different necessities. Regardless, most of them use dynamic controlling techniques to guide and move inside the line. Dependence upon the robot's course inside the line added to the troubles, in like way leaving the robot delicate against slip in the occasion that balance control procedures are not involved. The Theseus \cite{fujun2013modeling}, PipeTron \cite{choi2007pipe}, and PIRATE \cite{ravina2016kit} robot series use separate bits for driving and driven modules that are interconnected by different linkage types. Each sections change or shift the bearing for straightening out turns. Additionally, extraordinary intriguing controlling makes such robots subject to sensor data and huge appraisal.

Pipe climbing robots with three changed modules are to a great extent the more consistent and give better convenientce. Earlier proposed restricted line climbers \cite{vadapalli2019modular,suryavanshi2020omnidirectional,vadapalli2021modular} have utilized three driving tracks composed similarly to one another, as MRINSPECT series of robots. In such robots, to reasonably control the three tracks, their rates were pre-portrayed for the line turns. This addressed a fundamental for the robot to put together line reshapes definitively at a particular bearing standing apart from the pre-portrayed rates \cite{vadapalli2019modular,suryavanshi2020omnidirectional}. In certifiable applications, the robot's course shift enduring it experiences slip in the tracks during headway. This limitation can be addressed by using a latently worked transmission part to control the robot. MRINSPECT-VI \cite{chang2017development,litopic} uses a multi-crucial transmission stuff part to control the paces of the three modules. Notwithstanding, for the division of the main role and speed to the three modules, the course of action of the transmission is used. This perspective made the focal yield (Z) to turn speedier than the other two outcomes (X and Y), making yield Z successfully affected by slip \cite{chang2017development}. This is caused considering the way that the results of the transmission doesn't give comparable energy to the data. Other really proposed oversees veritable outcomes regarding transmissions \cite{kota1997systematic,kant2022pipe} additionally followed a comparative arrangement.

Transmission kills the suggested requirement by seeing obscure outcome to enter dynamic relations \cite{vadapalli2021design}. In the imagined arrangement all of the three outcomes are proportionately affected by the information. This contributes for the robot to crash slip and drag toward any method of the robot during its new development. Additionally, the transmission part in the line climber redesigns the comfort by decreasing the dependence on the solid controls to go through the line affiliations. The structure definitively moves power and speed from a lone commitment to the robot's three tracks through complex stuff trains, considering the stacks experienced by each track freely.
\section{Design}

A whimsical nonagon central bundling of the robot houses three modules isolated from each other. The transmission coordinated inside the central event of the robot drives the three tracks by its driving sprockets through. incline furnishes, that are associated with yields. The planned viewpoint on the instrument and is clear in the resulting subsection. Each module houses a track and has openings for the four linkages to slide. The modules are pushed radially outwards with the help of direct springs mounted on the linkages (or shafts) i.e., the springs between the modules and the robot body. The fitting is a reference structure that limits the headway of the modules past OK endpoints. Right when the robot is sent inside the line, the spring-stacked tracks go through pulled out evasion and presses against the line's internal dividers. It gives the chief balance to the robot to move. Each module in the robot can in like way pack unevenly, point by point partially. The tracks, obliged on the modules, are crushed against the inward dividers of the line which give the immense balance to the robot to move. There is a driving sprocket that changes over the rotational headway from the inevitable result of to translatory improvement for the robot. The divider crushing design contains 4 direct linkages for the three modules overall. The modules have openings through which the straight linkages (or shafts) pass. They grant the headway of the modules in the curving headings from the turn of the robot.

\vspace{-0.15in}
\begin{figure}[ht!]
\centering
\includegraphics[width=2.5in]{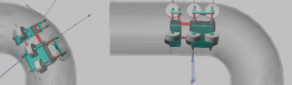}
\vspace{-0.15in}
\caption{\footnotesize Robot}
\label{1}
\vspace{-0.15in}
\end{figure}
The transmission is the significant constituent of the proposed robot. The instrument contains a singular data, three transmissions, three two-input open transmissions and three outcomes. The transmission's examination is coordinated at the central turn of the robot body. The three are formed reliably around the data, with a place of between any two. The are fitted radially in the center. The single aftereffect of improvement the three consequences of the .Includes gear parts, for instance, ring gears, incline pinion haggles machine gear-piece wheels, while join ring gears, incline pinion haggles pinion wheels. The side machine gear-piece wheels ofis concurred with their close by side pinion wheels of, to move the power and speed from to it's lining. The data of the worm gear give headway to simuntaneously. Each two-yield transmission then, trade got progression to its coating two-input transmissions , dependent upon the load conditions experienced by its different side pinion wheels. The improvement got by the side pinion wheels of extra makes a translation of them to the three outcomes. The six transmissions and work together to disentangle improvement from the commitment to the three outcomes. Right when experience different burdens, the side pinion wheels in trades different burdens to the side pinion wheels of. Under this condition, makes an understanding of transmission speed to its lining. 

Definitively when experience identical weight or no stack, all of the side pinion wheels experience an equivalent weight making them turn at an overall speed and power. As spread out, all of the outcomes share undefined energy with the information. Besides, the outcomes in like way share hazy energy with each other. This results in the separation in loads in one of the outcomes obliging a comparable effect on the other two outcomes that are undisturbed. The outcomes work with indistinct speeds when there is no store or comparative weight returning again to the outcomes overall. The part works its outcomes with transmission speed when the outcomes are under changed burdens. Definitively when one of the outcomes is working at a substitute speed while the other two outcomes are experiencing commensurate weight, then, the two outcomes with comparable burdens will work with undefined speeds. The part expected the robot, pushes its the three tracks with unclear speeds while moving inside a straight line. Regardless, while moving inside the line curves, the transmission changes the track speed of the robot with a conclusive objective that the track embarking to all pieces of the more broadened distance turns faster than the track embarking to all pieces of the more restricted distance. Suggest \cite{vadapalli2021modular,9635853,9517351} for additional figures, diagrams and data.
\section{Design Specifications}

The kinematic plot shows the transparency of the affiliations and joints of the instrument. The transmission's kinematic and dynamic conditions are coordinated using the bond chart model. The data exact speed from the motor is in like way provided for the three ring gear-tooth wheels of the two-yield transmissions as. They turn at unclear exact rates and with for all intents and purposes indistinguishable power for instance times is the stuff level of the commitment to the ring gears. Additionally, a two-yield transmission coordinates that the particular speed of its ring gear is the standard every time of the cheeky rates of its two side pinion wheels. These two side stuff tooth wheels can turn at different rates while staying aware of vague power \cite{deur2010modeling}.
\begin{equation}
\arctan(AOC) = \theta = 90^0,
\label{1}
\end{equation}
From the security chart model, we can interpret that the three ring gears turns at indistinct exact rates and with for all intents and purposes indistinguishable power for instance times the data running velocity, events the data power, is the stuff level of the commitment to the ring gears. Also, in a two-yield transmission, the particular speed of its ring gear is the customary all vital of the cheeky rates of its two side pinion wheels. These two side machine gear-piece wheels can turn at different rates while staying aware of indistinct power \cite{vadapalli2021design}.The testing rates are then surmised their side machine gear-piece wheels. Side machine gear-piece wheels are associated with a conclusive objective that they don't have any expansive improvement between the pinion wheels of a commensurate pair, i.e., exact paces of side pinion wheels of an identical pair is by and large vague. Subbing in (\ref{1}). Basically, the outcome precise rates. The outcome to side stuff association is wandered from achieve the cheerful speed condition for the commitment to the outcomes. The trying paces of the ring pinion wheels to the side machine gear-piece wheels, we achieve a relationship among information and side stuff tooth wheels. Likewise, the result careful not completely settled from the speeds of the solitary ring gear-tooth wheels to get a result to side stuff affiliation. Looking at both the relations, we achieve the specific speed condition for the obligation to the results. where is the particular speed of the data is the stuff level of the ring gear-tooth wheels to the outcome, while shrewd rates of the outcome. The individual exact rates of the side stuff tooth wheels. Along these lines, the outcome precise rates are dependent upon the data running rate ($\omega_u$) and the side stuff yields. Meanwhile, the powers of the ring gears is how much the powers of their seeing side pinion wheels. Like cautious speed relationship for commitment to yields, by separating the ring pinion wheels with side pinion wheels relationship with the outcome to the ring gear alliance, a connection between yield powers is gained.
\begin{equation}
AC = 28 mm
\label{2}
\end{equation}
where is the particular speed of the data is the stuff level of the ring machine gear-piece wheels to the outcomes, while and are running speeds of the outcome. Yields are the individual precise paces of the side stuff tooth wheels. In this manner, the outcome exact rates are dependent upon the data light speed and the side stuff yields. Meanwhile, the powers of the ring gears is how much the powers of their seeing side pinion wheels. Like exact speed relationship for commitment to yields, by separating the ring pinion wheels with side pinion wheels relationship with the outcome to the ring gear connection, a connection between yield powers to the data power is acquired.
\begin{equation}
MN > X^2 + Y^2 - Z^2
\label{3}
\end{equation}
\begin{equation}
MN < AC * \theta + Y^2
\label{4}
\end{equation}
Conditions \eqref{3} and \eqref{4} presents the interest of the transmission to give comparable progression ascribes in all of the three outcomes that experiences indistinct burdens or when left unconstrained. Notwithstanding, when the stuff parts experience an impediment across a convergence point, the smart rate and the power changes depending outwardly resistive power. The outcome speed of the transmission results are like the speed of the driving module sprockets.Therefore, the outcome paces of the transmission are changed more than into track speeds. The data speed for the robot is, thusly making an understanding of to the outcomes under essentially indistinguishable stacking conditions. The sprocket broadness is consistent for the three tracks overall. Ho Moon Kim et al. \cite{litopic}, in their paper proposed a method for working out the specific velocities of the three tracks inside pipe turns. Bearing that the robot enters the pipeline with the course of action showed is the attribute of the line turn, R is the extensiveness of melodic improvement of the line wind and r is the degree of the line. Furthermore, the paces for their specific tracks B and C are gotten. The robot is implanted at different heading of the modules concerning OD. The changed speeds for the not forever set up for wind pipes in headings.

\subsection{Constraints}

The straight springs in the module gives robot the versatility to orchestrate turns with fundamentally no issue. The best strain in each module is $16 mm$. There are additional versatilities in the module openings, with the genuine that upside down pressure is possible. This helps the robot with overcoming obstructions and idiosyncrasies in the line network that it may review genuine applications. The front fulfillment of the module is compacted absolutely in any case the back is in its most significant expanded state possible inside the line. The most senseless unequal strain allowed in a single module of the robot. Along these lines, $\phi$ is the best point the module can pack unevenly. Infer \cite{vadapalli2021modular,9635853,9517351} for additional figures, blueprints and data.
\vspace{-0.1in}
\begin{figure}[ht!]
\vspace{-0.1in}
\centering
\includegraphics[width=2.75in]{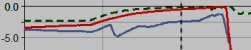}
\vspace{-0.12in}
\caption{\footnotesize Bends}
\label{asym(1)}
\vspace{-0.1in}
\end{figure}

\section{Experimentation}

Redirections were facilitated to check out and uphold the improvement farthest reaches of the robot in various line affiliations. An equivalent will give us more encounters into the parts and direct of the made robot in authentic testing conditions. From here on out, multi-body dynamic spreads was acted by shifting over the direction of activity limits into a redesigned reenactment model. To reduce how much moving parts in the model and to lessen the computational weight, the tracks were modified into roller wheels. Each module houses three roller wheels in the better model. Thusly, the contact fix given by the tracks to the line dividers are diminished from 10 contact locale to 3 contact regions for each module. Quite far favor the track speeds and the module strain for each track A,B and C were taken apart. Increments were performed by installing the robot in three irrefutable headings of the module in both the straight lines and line turns. The robot is reiterated inside a line network coordinated by ASME B16.9 standard NPS 11 and plan 40. The redirections were created for four evaluation conditions in the line network containing Vertical region, Elbow region, Horizontal piece and the U-part for different heading of the robot. The totally distance of the line structure. The distance went by the robot in not forever set up from point of blend of the robot body and the track's single distance actually hanging out there from the center roller wheel mounted in each module. In this manner, we get the without a doubt robot's course, by killing the robot's length.The robot's way in vertical climbing and the last level area is studied by deducting from their single piece length. The information of the is given a strong perfect speed and improvement of the robot including the track speeds are considered in the reenactment.

In the vertical piece and level region, the robot follows a straight way. Thus, the tracks experience vague burdens on all of the three modules in both the assessments. Hence, the transmission gives undefined rates to the three tracks as a general rule, similar to the robot's regular speed. The saw track speeds in the redirection for the heading. In like manner, all of the characteristics stand apart from the speculative results with a level out rate screw up (APE). This slip-up overviews the certifiable degree of deviation from the hypothetical worth \cite{armstrong1992error}. To assemble a critical length of in vertical climbing, the robot guesses that 0 should 9 seconds. the robot's ability to climb the line confronting gravity. Starting the robot moves a distance of 350 mm in the supervisor even area. In the seriously unassuming level piece of distance. In the elbow locale, the robot moves at a consistent distance to the spot of get together of twist of the line. The framework changes the outcome speeds of the tracks as shown by the package from the spot of get together of state of the line. In every one of the three headings of the robot, the outer module tracks goes speedier to travel a more broadened distance, while the inside module tracks turns considerably more delayed to embark to all pieces of the most restricted distance than the extent of touch of the line twist. In pipe reshapes, the redirection rates of each track is shown at the midpoint of seperately to ruthless the saw track speeds without changes. The approximated rates of each track is then stood separated from their specific hypothetical rates with track down the overall rate botch (APE).

For the course, the outside modules (B and C) move at a typical speed of, while the inside module (A) move at a standard speed of 33.62mm/sec. These characteristics interact with the speculative traits. Likewise, the track speeds organizes with the standard worth of the development results. In addition, the track speed a motivation, veer from the augmentation results with an APE of 2.5\%. Toward each method of both the straight and turn regions, the screw up regard is unfathomably unessential and they can be credited on account of outside factors in clear conditions like disintegrating. Likewise, unimportant speed changes occurs in the increase plot. From 9 to 24 seconds, the robot assembles the elbow area of distance, while it requires 59 seconds to go in the U-portion. The robot's ability to research in pipe-turns. for shows that the outer modules (B and C) move at the speed degree while within module (A) move in a speed degree. These credits exist in the hypothetical characteristics. Likewise, the track speeds a motivation, exist in the reenactment results. The speed range implied from the redirection is taken from the base to the most preposterous highest points of rates for each plots. Imply \cite{vadapalli2021modular,9635853,9517351} for additional figures, diagrams and data.

The extension results for the track speeds in different headings, arranges with the theoretical results got in area. It is seen from the reenactment that in 60 seconds, the robot researches all through the line network reliably at the installed bearing. The result connect with the hypothetical calculation. This upholds that the transmission discards slip and drag in the tracks of the line climber done with no improvement hardships. In the expansion, the robot is seen with no slip and drag everywhere, which further impacts in diminished tension effect on the robot and extended headway flawlessness. The track in the modules catch to the interior mass of the line to give balance during headway. The springs are at first pre-stacked by a strain in each of the three modules comparatively when inserted in the vertical line region. In straight lines, the robot moves at the fundamental pre-stacked spring length. The contortion length increases by 1.5 mm for inside and the outer modules when the robot is moving near elbow region and U-section. This deformation explains the extensive versatility pondered the modules to go through the making cross-part of the line broadness in the turns during progress.

\section{Conclusion}
The robot is given the splendid transmission to control the robot unequivocally with no extraordinary controls. The transmission has an unclear outcome to join energy, whose show is absolutely like the solace of the standard two outcome transmission. The age results support solid intersection purpose in staggering line networks with spots of up to 180$^\circ$ in different headings without slip. Taking on the transmission part in the robot achieves the sharp deferred aftereffect of getting out the slip and drag all over of the robot during the new development. At the present, we are engaging a model to perform investigates the proposed arrangement.

\bibliographystyle{asmems4}

\bibliography{asme2e}

\end{document}